\documentclass[wcp]{jmlr}

\usepackage{longtable}
\usepackage{graphicx}
\usepackage{amsmath}
\usepackage{algorithm}
\usepackage{algorithmic}
\usepackage{multirow}
\usepackage{amsfonts}
\usepackage{hyperref}
\newcommand\fnurl[2]{%
	\href{#2}{#1}\footnote{\url{#2}}%
}
\makeatletter
\renewcommand*{\thanks}[1]{%
	\footnotemark
	\protected@xdef\@thanks{\@thanks
		\protect\footnotetext[\arabic{footnote}]{#1}}%
}
\makeatother
\usepackage{booktabs}

\jmlrvolume{101}
\jmlryear{2019}
\jmlrworkshop{ACML 2019}

\title[self-paced multi-label learning with diversity]{Self-Paced Multi-Label Learning with Diversity}

 \author{\Name{Seyed Amjad Seyedi} \Email{amjadseyedi@eng.uok.ac.ir}\\
   \Name{S. Siamak Ghodsi}\thanks{S. A. Seyedi and S. S. Ghodsi---Equal Contribution.} \Email{s.ghodsi@eng.uok.ac.ir}\\
  \Name{Fardin Akhlaghian} \Email{f.akhlaghian@uok.ac.ir}\\
  \addr Department of Computer Engineering, University of Kurdistan, Sanandaj, Iran
  \AND
  \Name{Mahdi Jalili} \Email{mahdi.jalili@rmit.edu.au}\\
 \addr School of Engineering, RMIT University, Melbourne, Australia
  \AND
  \Name{Parham Moradi}\thanks{Corresponding Author} \Email{p.moradi@uok.ac.ir}\\
  \addr Department of Computer Engineering, University of Kurdistan, Sanandaj, Iran
 }

\editors{Wee Sun Lee and Taiji Suzuki}

\begin{document}

\maketitle

\begin{abstract}
	
The major challenge of learning from multi-label data has arisen from the overwhelming size of label space which makes this problem NP-hard. This problem can be alleviated by gradually involving easy to hard tags into the learning process. Besides, the utilization of a diversity maintenance approach avoids overfitting on a subset of easy labels. In this paper, we propose a self-paced multi-label learning with diversity (SPMLD) which aims to cover diverse labels with respect to its learning pace. In addition, the proposed framework is applied to an efficient correlation-based multi-label method. The non-convex objective function is optimized by an extension of the block coordinate descent algorithm. Empirical evaluations on real-world datasets with different dimensions of features and labels imply the effectiveness of the proposed predictive model.
\end{abstract}

\section{Introduction}

The paradigm of multi-label learning has become a popular topic in recent years. In many real-world applications, instances are in semantic association with more than one class label \cite{8423669}. Therefore, it sounds more rational to map each instance into a vector of labels rather than a single one. An effective stage to handle a multi-label problem is to learn dependencies among the labels \cite{zhang2014review}. Accordingly, numerous studies have been conducted to accomplish this goal. \cite{zhang2007ml} proposed a lazy learning method, derived from the conventional kNN classifier which classifies instances using a statistical model based on maximum a posteriori principle. However, this algorithm implicitly covers a local definition of correlation; label dependency has gone further.  \cite{huang2012multi} investigated an explicit view of local correlation by encoding their influences into a local code (LOC).

Another important stage that has drawn much attention is a low-rank representation of label space. Many algorithms with this property are proposed. \cite{yu2014large} presented a large-scale low-rank structure applicable to scaled label spaces meanwhile handling data with missing-labels. The framework of \cite{xu2014learning} aims to capture global label correlations by utilizing a low-rank structure on the label correlation matrix and copes with the missing-label challenge by introducing a supplementary label matrix. \cite{xu2013speedup} proposed a fast matrix completion algorithm with a low-rank representation which exploits side information explicitly to optimize complexity in a transductive manner.  \cite{8233207} investigated the concept of label correlation in a new manner. Contrary to previous approaches that only rely on a single definition of correlation i.e. global or local, the GLOCAL framework analyzes both GLObal and loCAL correlations of labels simultaneously in a latent label representation. This method takes advantage of manifold optimization and is capable of dealing with both missing and full label scenarios.

Algorithms mentioned so far have many pros and cons. Many methods in this field have studied the multi-label problem from different perspectives and have made valuable efforts. However, there is a common shortcoming;  they lack a mechanism to give a clear order to training instances. Some instances are easy for a specific label. It is beneficial to learn that label with those instances first and then gradually learn harder ones. For example, learning label "rabbit" with a black rabbit running on grass in a picture is easier than a white rabbit running on snow.

Curriculum and self-paced learning (SPL) are recently proposed regimes with the aim of learning from easier to more complex concepts \cite{bengio2009curriculum,kumar2010self,meng2017theoretical}. These learning frameworks are inspired by human education system where the major difference arises in identifying the complexity level. Curriculum learning needs a teacher (extra knowledge) to distinguish easy concepts from the complex ones, whereas self-paced learning is like a student who starts to learn a curriculum based on self-abilities.

The SPL framework is widely applied to various fields. \cite{zhao2015self} incorporated SPL with conventional matrix factorization and introduced a new Matrix factorization framework with a generalization of SPL to produce soft weight values along with the original binary weights. \cite{li2017self} proposed a multi-task algorithm with an self-paced regularization, and optimized this learner with efficient development of block coordinate descent.  Self-Paced learning is claimed to be a general framework applicable to any learning framework having an objective function with an empirical loss function. It has been successfully applied in various learning fields such as classification \cite{li2017selfcnn}, boost learning \cite{pi2016self}, object detection \cite{sangineto2019self}, Co-saliency detection \cite{zhang2017co}, face identification \cite{lin2018active}, Multi-view Clustering \cite{xu2015multi}, and multi-task learning \cite{murugesan2017self}.

\cite{li2018self} introduced a self-paced regularization framework for multi-label learning. It is one of the first attempts to tackle the multi-label problem by considering the complexity of instances for labels. However, without a diversity maintenance approach, a self-paced regularizer may cause the learning model to be biased on a subset of labels that are easy to learn \cite{jiang2014self}.

In this paper, a self-paced multi-label with diversity (SPMLD) framework is proposed. The diversity regularization term drives the model to be inclined to learn different labels that are easier first and somehow overcomes the problem of being biased on a limited number of easy labels. Besides, this gradual learning scheme can exploit more reliable label dependencies. Finally, to present a comparable example for realizing desired self-paced multi-label learning, SPMLD is applied to a host algorithm, a recent multi-label method \cite{8233207} which has acceptable performance. Empirical results supported by statistical significance tests demonstrate the effectiveness of our method against several well-known algorithms including the host algorithm.

\section{Background}
Self-paced learning provides a way for simultaneously choosing the easier patterns and re-estimating the learning parameters $\textbf{w}$ in the form of an iterative process \cite{kumar2010self}.
We presume a linear function $\mathit{f}(\textbf{\emph{x}}_i,\textbf{{w}})$ with unknown parameter \textbf{w}. SPL is then given by the following objective function to be solved:
\begin{align}\label{f1}
\min_{\textbf{{w,p}}\in \Omega} & \sum_{i=1}^n p_i \ell_i(y_i,\mathit{f}(\textbf{\emph{x}}_i,\textbf{{w}}))
+ \Psi(\lambda,\textbf{p})
\end{align}
where $\Psi(\lambda,\textbf{p})=\lambda\sum_{i=1}^{n}p_i$ is the regularization term, $\Omega$ is the domain space 
of $\textbf{p}$ and $\Psi(\lambda,\textbf{p})$ is the regularization parameter
which determines the complexity of patterns.
Equation (\ref{f1}) has two unknowns including $\textbf{w}$ as learning parameter and $\lambda$ the parameter of pace control (restricted to the specified domain).
Equation (\ref{f1}) then becomes a bi-convex optimization problem over the parameters \textbf{w} and $\textbf{p}$, which can be efficiently solved by alternating minimization. The optimal solution of $\textbf{w}$ with fixed $\textbf{p}$ can be achieved by any off-the-shelf solver and,
the optimal solution of $\textbf{p}$ with a fixed $\textbf{w}$ can be obtained by:
\begin{align}\label{f2}
p^{\ast}_{i}=
\begin{cases}
\begin{aligned}[b]
1,& \end{aligned} &\ell_i(y_i,\mathit{f}(\textbf{\emph{x}}_i,\textbf{{w}}))<\lambda,\\
0 ,&\mathrm{otherwise.}\\
\end{cases}
\end{align}
According to (\ref{f2}) easy samples have losses smaller than a determined threshold because they have less prediction errors, when updating $\textbf{p}$ given a fixed \textbf{w}, so they are chosen for training ($p^{\ast}_{i}=1$) or otherwise they aren't chosen ($p^{\ast}_{i}=0$). for updating \textbf{w} given a fixed $\textbf{p}$, the training process of learning model only performs on the "easy" samples selected before. Small values of $\lambda$, only pass "easy" samples with small losses. With gradually increasing $\lambda$, larger loss values for "complex" samples are accepted.
\section{Self-Paced Multi-Label Learning}
Suppose we are given an input matrix $\textbf{X} \in \mathbb{R} ^{d\times n}$ with \textit{d}-dimensional feature space and $\textbf{Y} \in\{-1,+1\} ^{l\times n}$
be the finite set of \textit{l} existing labels. Let  $\mathcal{D} = {(\textbf{x}_j, \textbf{y}_j)}^{n}_{j=1}$ be a multi-label
data, where $\textbf{x}_j = [x_{1j},..., x_{dj}]$ is an arbitrary vector of features with \textit{d}-dimensions of the \textit{j}th sample and $\textbf{y}_j = [y_{1j},..., y_{lj}]$ is the interpretation of labels for $\textbf{x}_j$ and $y_{ij}$ is $+1$ if the \textit{i}th label is assigned to $\textbf{x}_j$ and $-1$
otherwise. Multi-label learning intends to train a predictor $\mathcal{H} : \mathcal{X}\rightarrow \mathcal{Y}$ from the training data $\mathcal{D}$, so relevant and irrelevant labels of out-of-sample data are predicted. The general objective function for multi-label learning is: 
\begin{align}
\min_{\textbf{W}} \sum_{i=1}^{l}\sum_{j=1}^{n}\mathcal{L}(y_{ij},\textbf{w}_i^{\top}\textbf{x}_j)+\Phi(\textbf{W}, \textbf{C})
\end{align}
where $\textbf{W}=\{\textbf{w}_1, ..., \textbf{w}_{l} \}\in \mathcal{R}^{d\times l}$ 
is the learned matrix with each column indicating the weight vector for each independent label. $\mathcal{L}(y_{ij},\textbf{w}_i^{\top}\textbf{x}_j)$ is the loss of \textit{j}th sample for the \textit{i}th label. Correlation matrix $\textbf{C}$ demonstrates dependency degree between each pair of labels and $\Phi(\textbf{W}, \textbf{C})$ is a correlation regularizer with characteristic to let labels with positive dependency degrees encourage their corresponding outputs to be closer, and vice versa.

The aforementioned objective function treats all labels identically and does similar for samples per label. Although in many real-world cases, different labels don't necessarily have identical complexities, and also samples have different complexity levels for labels. Generally, the non-convex objective function of multi-label learning is the potential to get stuck in local optima \cite{zhao2015self}, especially with the presence of bad initialization or noisy and corrupted labels.
To address these defects, by defining a self-paced regularization, the model can learn a sequence of instances with respect to their degree of complexity.
\begin{align}\label{ml1}
\min_{\textbf{W, P}}& \sum_{i=1}^{l}\sum_{j=1}^{n}p_{ij}\mathcal{L}(y_{ij},\textbf{w}_i^{\top}\textbf{x}_j)+\Phi(\textbf{W}, \textbf{C})+ \Psi(\lambda, \gamma,\textbf{P})\nonumber\\ &
\text{s.t.\phantom{s.t.}}\textbf{P} \in{[0,1]^{l\times n}},
\end{align}
Furthermore, our desirable multi-label learning is expected to learn not only easy but also diverse labels that are sufficiently disparate from the current learning pace. To this end, instance-label weights are introduced. In order to accomplish the easy-to-hard strategy on diverse labels simultaneously, we propose a new self-paced regularizer in \eqref{reg1}:
\begin{align}\label{reg1}
\Psi(\lambda, \gamma,\textbf{P})=-\lambda\sum_{i=1}^{l}\sum_{j=1}^{n}p_{ij}
+ \gamma\sum_{i=1}^{l}\begin{Vmatrix}\textbf{{p}}^{(i)}\end{Vmatrix}_{2}
\end{align}
Equation \eqref{reg1} consists of two terms, a negative $\textit{l}_1$-norm and an adaptive $l_{2,1}$-norm of a matrix. The first term induces a preference to select the easy instances rather than
the hard ones per label. Combining this term with \eqref{ml1}, implies that small empirical loss $\mathcal{L}$ on the training data
point ($\textbf{x}_j, y_{ij}$) drives the weight $p_{ij}$ to be high.
Hence, this optimization process well corresponds with the intuitive notion of
starting with the easiest instances (the ones that have the lowest empirical
errors). By progressively increasing $\lambda$  while the learning
proceeds, the self-paced weights will increasingly grow higher in consequent. This leads to gradual involvement of more complex samples into training. The $l_{2,1}$-matrix norm term leads to label-wise sparsity. It favors selecting from different categories of labels in the initial steps with higher diversities. 
As the training proceeds, with gradually decreasing $\gamma$, the impact of diversity decreases.  
By plugging \eqref{reg1} into \eqref{ml1}, we obtain the final objective function:
\begin{align}
\min_{\textbf{W, P}}& \sum_{i=1}^{l}\sum_{j=1}^{n}p_{ij}\mathcal{L}(y_{ij},\textbf{w}_i^{\top}\textbf{x}_j)+\Phi(\textbf{W}, \textbf{C})-\lambda\sum_{i=1}^{l}\sum_{j=1}^{n}p_{ij}
+ \gamma\sum_{i=1}^{l}\begin{Vmatrix}\textbf{{p}}^{(i)}\end{Vmatrix}_{2}\nonumber\\
& \text{s.t.}\;\textbf{p}^{(i)} \in{[0,1]^{n}}
\end{align}
\subsection{SPMLD with Local and Global Correlation}
We give an example to motivate our self-paced learning procedure. We briefly discuss how the proposed algorithm develops the learning pace of the original problem. Host method (GLOCAL) \cite{8233207} simultaneously recovers the missing-labels, trains the learner and exploits both global and local correlations among labels without needing any further prior knowledge, through learning a latent label representation. The objective function of GLOCAL is as follows:
\begin{align}
\min_{\textbf{{U,V,W,Z}}} & \begin{Vmatrix}\textbf{{J}} \circ (\textbf{{Y}}-\textbf{{UV}})\end{Vmatrix}_F^2 
+ \alpha\begin{Vmatrix} (\textbf{{V}}-\textbf{{W}}^{\top}\textbf{{X}})\end{Vmatrix}_F^2\nonumber\\&
+\sum_{b=1}^g [ \frac{\beta_1 n_b}{n} \mathrm{tr}(\textbf{{F}}^{\top}\textbf{{Z}}_b\textbf{{Z}}_b^{\top}\textbf{{F}})
+{\beta_2} \mathrm{tr}(\textbf{{F}}_b^{\top}\textbf{{Z}}_b\textbf{{Z}}_b^{\top}\textbf{{F}}_b)]\nonumber\\&
\text{s.t.}\phantom{s.t.} \:  \mathrm{diag}(\textbf{{Z}}_b\textbf{{Z}}_b^{\top}) = \textbf{{1}},\: b = 1,2,.. .,g.
\end{align}
where \textbf{V} stands for the matrix of the latent labels capturing concepts in a higher level which are more compact and semantically abstract than the original labels; while \textbf{U} represents the matrix containing the interactions between the original labels and the latent labels.
In general, labels may only be partially observed. Low-rank representation is one of the key techniques in matrix completion, and the low-rank decomposition of the observed labels yields a natural solution to recover missing-labels (Let \textbf{J} be the indicator matrix of the observed labels in \textbf{Y}).

Label correlations may not have the same values in different categories, so we define the local manifold regularization. Assume that the dataset \textbf{X} is partitioned into \textit{b} groups $\{\textbf{X}_1,…,\textbf{X}_b\}$, where $\textbf{X}_b \in \mathbb{R}^{d\times n_b}$ has $n_b$ instances.
This partitioning can be obtained by clustering. If $\textbf{Y}_b$ is the label submatrix in \textbf{Y} corresponding to $\textbf{X}_b$, then $\textbf{C}_b \in \mathbb{R}^{ l\times l}$ are the local correlation of group \textit{b}. Similar to global label correlations, we force the output to be similar or dissimilar on the relevant or irrelevant correlated labels, and optimize $\mathrm{tr}(\textbf{F}_b^{\top}\textbf{ L}_b \textbf{F}_b)$, where $\textbf{L}_b=\textbf{{Z}}_b\textbf{{Z}}_b^{\top}$ is the Laplacian of $\textbf{C}_b$
that stores our knowledge about the relationship of our labels
and is defined as $\textbf{{L}}_b=\textbf{{D}}_b-\textbf{{C}}_b$ where $\textbf{D}_b$ is a diagonal matrix with $d_{i,i}=\sum_{j=1}^n c_{i,j}$
and $\textbf{F}_b = \textbf{UW}^{\top} \textbf{X}_b$ is the predicted submatrix for group \textit{b}.\\
Finally, We propose the following objective function by considering the diversity of labels in a
unified setting:

\begin{align}\label{eq8}
\min_{\textbf{{U,V,W,Z,P}}} & \begin{Vmatrix}\textbf{{J}} \circ (\textbf{{Y}}-\textbf{{UV}})\end{Vmatrix}_F^2 
+ \alpha\begin{Vmatrix}\sqrt{\textbf{{P}}} \circ (\textbf{{V}}-\textbf{{W}}^{\top}\textbf{{X}})\end{Vmatrix}_F^2\nonumber\\&
+\sum_{b=1}^g [ \frac{\beta_1 n_b}{n} \mathrm{tr}(\textbf{{F}}^{\top}\textbf{{Z}}_b\textbf{{Z}}_b^{\top}\textbf{{F}})
+{\beta_2} \mathrm{tr}(\textbf{{F}}_b^{\top}\textbf{{Z}}_b\textbf{{Z}}_b^{\top}\textbf{{F}}_b)]\nonumber\\&
-\lambda\sum_{i=1}^l\begin{Vmatrix}\textbf{{P}}^{(i)}\end{Vmatrix}_1
+ \gamma\begin{Vmatrix}\textbf{{P}}^{\top}\end{Vmatrix}_{2,1}+\tau\mathcal{R}(\textbf{{U}},\textbf{{V}},\textbf{{W}})  \nonumber\\
\text{s.t.}\phantom{s.t.} &\:   \textbf{{P}} \in{[0,1]^{l \times n}},\mathrm{diag}(\textbf{{Z}}_b\textbf{{Z}}_b^{\top}) = \textbf{{1}},\: b = 1,2,.. .,g.
\end{align}
where $\sqrt{\textbf{{P}}}$ denotes
the element-wise square root of \textbf{P}, the $ \circ$ (element-wise product) of matrices and $\mathcal{R}(\textbf{{U}},\textbf{{V}},\textbf{{W}})=\begin{Vmatrix}\textbf{{U}} \end{Vmatrix}_F^2
+\begin{Vmatrix}\textbf{{V}} \end{Vmatrix}_F^2
+\begin{Vmatrix}\textbf{{W}} \end{Vmatrix}^2_{F}$ 
is the regularization term to guarantee generalization ability and numerical stability.
\begin{algorithm}[tb]
	\caption{SPMLD on GLOCAL}
	\label{alg:algorithm}
	\textbf{Input}: data matrix $\textbf{{X}}$, label matrix $\textbf{{Y}}$, Observation indicator matrix $\textbf{{J}}$, and the group partition\\
	\textbf{Parameter}: $\alpha,\beta_1,\beta_2,\tau,\lambda, \gamma$\\
	\textbf{Output}: $\textbf{{U}}$ and $\textbf{{W}}$
	\begin{algorithmic}[1] 
		\STATE Initialize $\textbf{{U}},\textbf{{V}},\textbf{{W}},\textbf{{Z}}$;
		\WHILE{convergence not reached}
		\FOR{$b = 1,2,.. .,g.$} 
		\STATE {fix\:$\textbf{{P}},\textbf{{U}},\textbf{{V}},\textbf{{W}}$, update\:$\textbf{{Z}}_b$\:according to   \eqref{zq};} 
		\ENDFOR
		\STATE {fix\:$\textbf{{P}},\textbf{{U}},\textbf{{W}},\textbf{{Z}}$, update\:$\textbf{{V}}$\:according to  \eqref{vq};} 
		\STATE {fix\:$\textbf{{P}},\textbf{{V}},\textbf{{W}},\textbf{{Z}}$, update\:$\textbf{{U}}$\:according to   \eqref{uq};} 
		\STATE {fix\:$\textbf{{P}},\textbf{{U}},\textbf{{V}},\textbf{{Z}}$, update\:$\textbf{{W}}$\:according to   \eqref{wq};} 
		\STATE {fix\:$\textbf{{U}},\textbf{{V}},\textbf{{W}},\textbf{{Z}}$, update\:$\textbf{{P}}$\: according to \eqref{myeq:one};} 
		\STATE $\lambda=\lambda\mu_1; \gamma=\gamma\mu_2;$
		\ENDWHILE ;
		\STATE \textbf{return} $\textbf{{U}}$ and $\textbf{{W}}$.
	\end{algorithmic}
\end{algorithm}

The details of self-paced multi-label learning with diversity (SPMLD) is summarized in Algorithm \ref{alg:algorithm}. 
Implementation is available on GitHub repository\fnurl{}{http://github.com/amjadseyedi/SPMLD}.
\subsection{Optimization}
In this section, we discuss how to solve (\ref{eq8}) by alternating minimization which gives us capability to tune the variables iteratively and find an optimal solution. It is difficult to find a global optimal answer for this non-convex objective function. To achieve the reliable self-paced weights, we extend a block coordinate descent optimizer. For solving block $\textbf{p}_{t+1}$ with fixed blocks $\textbf{U}_{t}$,$\textbf{V}_{t}$,$\textbf{W}_{t}$ and $\textbf{Z}_{t}$, the optimization problem can be decomposed to \textit{k} sub-problems for\textit{ k} latent labels, respectively. Thus, objective function of the \textit{i}-th label, $\textbf{y}_{i}$ is given by:
\begin{align}\label{eq:2.1}
\min_{\textbf{p}^{(i)}}  \textbf{p}^{(i)}\mathcal{L}^{(i)}_{t}-\lambda\begin{Vmatrix}\textbf{{p}}^{(i)}\end{Vmatrix}_1
+ \gamma\begin{Vmatrix}\textbf{{p}}^{(i)}\end{Vmatrix}_{2}
,\:\text{s.t.}\;\textbf{p}^{(i)} \in{[0,1]^{n}},
\end{align}
In order to solve \eqref{eq:2.1}, we first assume $\mathcal{L}^{(i)}_{1,t}\leq\mathcal{L}^{(i)}_{2,t}\leq...\leq\mathcal{L}^{(i)}_{n,t}$. Let $r^{(i)}_{t}=\sum_{\theta_1<j<\theta_2}(\lambda-\mathcal{L}^{(i)}_{j,t})^2$ and $s^{(i)}_{t}=\sum_{\theta_1<j<\theta_2}(\lambda-\mathcal{L}^{(i)}_{j,t})$. For each \textit{i} and arbitrary $\theta_2>\theta_1$%
we define $c_{t}^\ast(\theta_1,\theta_2)$, $L_{t}(\theta_1,\theta_2)$, $G_{i,t}^\ast$ and $H_{i,t}^\ast$ for later computation:
\\
\\ \textbf{1.}
\begin{align}
c_{t}^\ast(\theta_1,\theta_2)=  
\begin{cases}
\begin{aligned}[b]
\sqrt{\theta_1\slash (\gamma^2-r^{(i)}_{t})} & \end{aligned} &,\gamma^2\neq r^{(i)}_{t}\\
(\lambda-\mathcal{L}^{(i)}_{\theta_1+1}) &,\gamma^2= r^{(i)}_{t},\gamma^2< s^{(i)}_{t}\\
0 &,\gamma^2= r^{(i)}_{t},\gamma^2\geq s^{(i)}_{t}.
\end{cases}\nonumber
\end{align}
\textbf{2.}
\begin{align}
L_{t}(\theta_1,\theta_2)=& \sum_{j=1}^{\theta_1}\mathcal{L}^{(i)}_{j,t}-\lambda(\theta_1+c_{t}^\ast(\theta_1,\theta_2)s^{(i)}_{t})+\gamma\sqrt{\theta_1+c_{t}^\ast(\theta_1,\theta_2)^2r^{(i)}_{t})}.\nonumber
\end{align}
\\
\textbf{3.} $G_{i,t}^\ast$ be the smallest \textit{j} such that $\mathcal{L}^{(i)}_{j,t}\geq\lambda$.%
\\
\textbf{4.} $H_{i,t}^\ast$ be the largest \textit{j} such that $\mathcal{L}^{(i)}_{j,t}\leq\lambda-\gamma$.\\%
\\Let $\theta_2=G_{i,t}^\ast$, and $\theta_1$ be obtained by optimizing the following objective function:%
\begin{align}
\theta_1=\arg\min_{H_{i,t}^\ast\leq\theta_1 <\theta_1}  L_{t}(\theta_1,\theta_2)\nonumber
\end{align}
Then, the optimal $\textbf{p}^{(i)}_{t+1}$ is given by
\begin{align}\label{myeq:one}
p^{(i)}_{j,t+1}=
\begin{cases}
\begin{aligned}[b]
1,& \end{aligned} &j\leq\theta_1\\
0 ,&j\geq\theta_2\\
c_{t}^\ast(\theta_1,\theta_2)(\lambda-\mathcal{L}^{(i)}_{j,t}) ,&\theta_1<j<\theta_2
\end{cases}
\end{align}
Then, we discuss update procedures for \textbf{U}, \textbf{V}, \textbf{W}, and \textbf{Z}. To optimize these variables with the gradient descent method, we utilize the MANOPT toolbox\fnurl{}{http://www.manopt.org} \cite{boumal2014manopt} with line search on the Euclidian and manifold spaces. \\\\
\textbf{Updating $\textbf{Z}_b$'s}: With \textbf{U}, \textbf{V}, \textbf{W} and $\textbf{p}_k$’s fixed:
for each $b\in \{1,...,g\}$. Due to the constraint
$\mathrm{diag}(\textbf{{Z}}_b\textbf{{Z}}_b^{\top}) = \textbf{{1}}
$, it has no closed-form solution, and we
solve it with projected gradient descent. The gradient of the
objective w.r.t. $\textbf{Z}_{b}$ is
\begin{align}\label{zq}
\nabla_{\textbf{{Z}}_b}=&\frac{\beta_1 n_b}{n}\textbf{{UW}}^{\top}\textbf{{X}}\textbf{{X}}^{\top}\textbf{{W}}\textbf{{U}}^{\top}\textbf{{Z}}_b
+\beta_2\textbf{{UW}}^{\top}\textbf{{X}}_b\textbf{{X}}_b^{\top}\textbf{{W}}\textbf{{U}}^{\top}\textbf{{Z}}_b
\end{align}
To satisfy the constraint $\mathrm{diag}(\textbf{{Z}}_b\textbf{{Z}}_b^{\top}) = \textbf{{1}}$, we project each
row of $\textbf{Z}_b$ onto the unit norm ball $\textbf{{z}}_{b,j,:}\leftarrow\textbf{{z}}_{b,j,:}/ \Vert \textbf{{z}}_{b,j,:}\Vert_2$ after each update. where $\textbf{{z}}_{b,j,:}$ is the \textit{j}th row of $\textbf{Z}_b$. \\\\
\textbf{Updating \textbf{V}:} With \textbf{U}, \textbf{W}, $\textbf{Z}_b$’s and $\textbf{p}_k$’s fixed: The gradient of the objective in \eqref{eq8} w.r.t. \textbf{V} is
\begin{align}\label{vq}
\nabla_{\textbf{{V}}}=&\textbf{{U}}^{\top}(\textbf{{J}} \circ (\textbf{{UV}}-\textbf{{Y}})) +(\textbf{{P}} \circ (\textbf{{V}}-\textbf{{W}}^{\top}\textbf{{X}}))+\tau\textbf{{V}}
\end{align}
\textbf{Updating \textbf{U}:} With \textbf{V}, \textbf{W}, $\textbf{z}_b$’s and $\textbf{p}_k$’s fixed:
Again, we use gradient descent and the gradient w.r.t. \textbf{U} is:
\begin{align}\label{uq}
\nabla_{\textbf{{U}}}=&(\textbf{{J}} \circ (\textbf{{UV}}-\textbf{{Y}}))\textbf{{V}}^{\top} +\tau\textbf{{U}}
+\sum_{b=1}^g\textbf{{Z}}_b\textbf{{Z}}_b^{\top}\textbf{{U}} [ \frac{\beta_1 n_b}{n} \textbf{{W}}^{\top}\textbf{{X}}\textbf{{X}}^{\top}\textbf{{W}}
+{\beta_2}  \textbf{{W}}^{\top}\textbf{{X}}_b\textbf{{X}}_b^{\top}\textbf{{W}}]
\end{align}
\textbf{Updating \textbf{W}:} With \textbf{U}, \textbf{V}, $\textbf{z}_b$’s and $\textbf{p}_k$’s fixed:
The gradient w.r.t. \textbf{W} is:
\begin{align}\label{wq}
\nabla_{\textbf{{W}}}
&=\alpha\textbf{{X}}[(\textbf{{X}}^{\top}\textbf{{W}}-\textbf{{V}}^{\top})\circ \textbf{{P}}^{\top}] +\tau\textbf{{W}}+\sum_{b=1}^g (\frac{\beta_1 n_b}{n}\textbf{{X}}\textbf{{X}}^{\top}+\beta_2  \textbf{{X}}_b \textbf{{X}}_b^{\top})
\textbf{{WU}}^{\top}\textbf{{Z}}_b\textbf{{Z}}_b^{\top}\textbf{{U}}
\end{align}
\section{Experiments}
In this section, empirical experiments are conducted to test the validation of our method. In these experiments, seven real-world multi-label datasets including Yahoo text datasets (\textit{Business, Computers, Education, Health, Science and Social}) \cite{ueda2003parametric} along with an image classification data (Corel5K) \cite{duygulu2002object} are used\fnurl{}{http://mulan.sourceforge.net/datasets-mlc.html}.

\subsection{Experimental Setting}

\begin{table}
	\caption{Statistical characteristics of the real-world multi-label datasets.}\label{tab1}
	
	\begin{tabular*}{\textwidth}{@{\extracolsep{\fill} }lcccc}  
		\hline
		dataset  & \#instance &  \#dimension &  \#label & \#label$\slash$instance\\
		\hline
		Business  \cite{ueda2003parametric}       & 5,000   & 438  & 30    & 1.59 \\
		Computers \cite{ueda2003parametric}       & 5,000  & 681  & 33    & 1.5 \\
		Education \cite{ueda2003parametric}       & 5,000  & 550  & 33    & 1.46 \\
		Health \cite{ueda2003parametric}      & 5,000  & 612  & 32    & 1.66 \\
		Science \cite{ueda2003parametric}      & 5,000  & 743  & 40    & 1.45 \\
		Social \cite{ueda2003parametric}      & 5,000  & 1,047  & 39    & 1.28 \\
		
		Corel5K \cite{duygulu2002object}       & 5,000  & 499  & 374    & 3.52 \\
		\hline
	\end{tabular*}
\end{table}

 Table 1 lists detailed characteristics of the employed datasets. Each column sequentially represents the number of features, number of instances, number of labels and label per instance ratio for each dataset. 
 
Since prediction in the presence of missing-labels is a more challenging task, we have performed the experiments on missing-label data. We randomly sample $\rho\%$ of the elements in the label matrix as observed, and the rest as missing. $\rho$ is set to 30\% and 70\% revealed entries, respectively. Coverage, Ranking loss, Average AUC, Instance AUC, MacroF1, MicroF1, and InstanceF1 evaluation metrics are used to measure the performance of the proposed predictive model against baselines. The above-mentioned metrics analyze different aspects of multi-label learning algorithms. The first two metrics are to be minimized and the other metrics are to be maximized according to \cite{wu2017unified}.

For examining the effectiveness of our framework, it is compared with three state-of-the-art multi-label learning algorithms namely LEML \cite{yu2014large}, ML-LRC \cite{xu2014learning}, GLOCAL \cite{8233207}. These Baseline methods along with the SPMLD have two common traits, which are the main reasons that we compare our framework with them. All the above-mentioned methods somehow learn in a latent subspace and furthermore, they all handle the missing-label challenge.

\begin{itemize}
	\item Low-rank empirical risk minimization for multi-label learning (LEML) has a linear low-rank structure to train a model for mapping from instance to label space and utilizes an implicit concept of global label dependency. 
	\item	Learning low-rank label correlations for multi-label
	classification (ML-LRC) learns and exploits low-rank global label correlations for multi-label classification.
	\item Multi-label learning with global and local label correlation (GLOCAL) learns label correlations in a new manner. It considers both local and global label correlations in latent label representation.  
\end{itemize}
What's more, to statistically measure the significance of performance difference, pairwise t-tests at 95\% significance level are conducted between SPMLD and each of the baseline algorithm. Therefore, in the statistical peer test, for each test, the performance of an algorithm is bold-faced denoting that it statistically outperforms the other one. Furthermore, when there is no significant difference between the performance of SPMLD compared to one or more of the baselines on a dataset regarding a specific evaluation metric, their results are shown in italic. Furthermore, self-paced regularization parameter $\lambda$ controls the complexity of instances and labels. It first considers the least losses corresponding to the easiest samples, then gradually involves harder ones through iterations. $\lambda$ and its increase ratio $\mu_1$ are searched from $\{10^{-1}, 10^{-2}, 10^{-3}, 10^{-4}, 10^{-5}\}$ and $\{1.1, 1.2, 1.3, 1.4, 1.5\}$, respectively. Diversity regularization parameter $\gamma$ controls to select less similar instances and labels specified as easy. $\gamma$ needs to be higher in the initial iterations. $\gamma$ and its decrease ratio $\mu_2$ are tuned using a grid search in ranges [1-10] and $\{0.95, 0.9, 0.85,0.8,0.75,0.7\}$, respectively. Parameters of the host algorithm (GLOCAL) and parameters of the competing methods are all set the same as recommended in the corresponding literature.

\subsection{Results on Real-World Datasets}
All the results reported in tables and charts are averaged over 10 independent runs. Tables (2-7) exhibit the obtained label prediction results of the multi-label algorithms on six of the mentioned datasets, respectively. Overall, results of all algorithms have improved on 70\% observation of samples. In each table three metrics are reported, regarding two different settings of $\rho$ there are six probable cases. On five datasets; \textit{Business}, \textit{Education}, \textit{Health}, \textit{Science} and \textit{Social}, SPMLD has significantly better performance regarding all the measures (and for both observation settings) reported in Tables (2, 4-6), respectively. On \textit{Computers} dataset whose results are shown in Table 3. SPMLD reports significantly better results regarding Rkl measure for both $\rho$ settings. Similarly, it shows statistically better AUC and COV values for 30\% entries revealed, while in the case of 70\% LEML, ML-LRC and SPMLD show no significant different AUC values compared to each other but they have jointly better AUC values than the GLOCAL. Again, in the case of COV values for 70\% entries, the proposed method shows no significant difference compared with ML-LRC but, it gains statistically better results than the two other models. Equivalently, on \textit{Social} dataset SPMLD shows significantly better performance regarding both percentages of observation for Rkl and AUC metrics and also COV with $\rho$=30\% and only for the proportion of 70\% it statistically performs equal to GLOCAL while they get jointly better results than the remaining two algorithms and jointly stand at the first place. Subsequently, as the summary of tables, SPMLD ranks first in 91.66\% cases according to statistical significance test. Moreover, it shows statistically equal performance in 8.34\% cases where it still ranks first but jointly with one ore more of the baselines.
\begin{table}
	\caption{Average performance on \textit{Business} dataset: means and standard deviations over
		10 independent runs. The best performance is highlighted in\textbf{ bold} with respect to
		paired t-tests at 95\% significance level between SPMLD and each baseline}
	\begin{tabular*}{\textwidth}{@{\extracolsep{\fill} }lccccc}  
		\hline
		Method &$\rho$   & Rkl $\downarrow$& AUC $\uparrow$ &COV $\downarrow$  \\ 
		\hline
		\multirow{2}{*}{LEML}
		&30\%  & 0.063 $\pm$ 0.0057 & 0.928 $\pm$ 0.0052 & 3.954 $\pm$ 0.2751   \\
		&70\%  & 0.058 $\pm$ 0.0048 & 0.942 $\pm$ 0.0057 & 3.303 $\pm$ 0.2706  \\ \hline
		\multirow{2}{*}{ML-LRC}
		&30\%  & 0.061 $\pm$ 0.0024 & 0.937 $\pm$ 0.0055 & 3.279  $\pm$ 0.0666  \\
		&70\%  & 0.046 $\pm$ 0.0019 & 0.950  $\pm$ 0.0050 & 2.580 $\pm$ 0.0593  \\ \hline
		\multirow{2}{*}{GLOCAL}
		&30\%  & 0.054 $\pm$ 0.0025 & 0.937 $\pm$ 0.0036 & 2.863 $\pm$ 0.1711   \\
		&70\%  & 0.046 $\pm$ 0.0022  & 0.952 $\pm$ 0.0031 & 2.579 $\pm$ 0.1658  \\ \hline
		\multirow{2}{*}{SPMLD} 
		&30\%  &\textbf{0.044 $\pm$ 0.0021}  & \textbf{0.956 $\pm$ 0.0032} & \textbf{2.361$\pm$ 0.1688}    \\
		&70\%  &\textbf{0.043 $\pm$ 0.0018}  & \textbf{0.958 $\pm$ 0.0029}  &\textbf{2.347$\pm$ 0.1625}   \\ 
		\hline
	\end{tabular*}
	
\end{table}
\begin{table}
	\caption{Average performance on \textit{Computers} dataset: means and standard deviations over 10 independent runs. The best performance is highlighted in \textbf{bold} with respect to paired t-tests at 95\% significance level between SPMLD and each baseline. \textit{Italic} font indicates that SPMLD and the corresponding baselines show no significant difference.}
	\begin{tabular*}{\textwidth}{@{\extracolsep{\fill} }lccccc}  
		\hline
		Method &$\rho$   & Rkl $\downarrow$& AUC $\uparrow$ &COV $\downarrow$  \\ 
		\hline
			\multirow{2}{*}{LEML}
	
	&30\%  & 0.179 $\pm$ 0.0072  & 0.880 $\pm$ 0.0064 & 7.392 $\pm$ 0.2181  \\
	&70\%  & 0.141 $\pm$ 0.0058  & \textit{0.894 $\pm$ 0.0069} & 6.306 $\pm$ 0.2653  \\ \hline
	
	\multirow{2}{*}{ML-LRC}
	
	&30\%  & 0.152 $\pm$ 0.0044  & 0.873 $\pm$ 0.0057 & 6.052 $\pm$ 0.1426  \\
	&70\%  & 0.115 $\pm$ 0.0026  & \textit{0.895 $\pm$ 0.0058} & \textit{5.000 $\pm$ 0.6228} \\ \hline
	
	\multirow{2}{*}{GLOCAL}
	
	&30\%  & 0.132 $\pm$ 0.0037  & 0.876 $\pm$ 0.0034 & 5.647 $\pm$ 0.7823  \\
	&70\%  & 0.123 $\pm$ 0.0028  & 0.884 $\pm$ 0.0062 & 5.440 $\pm$ 0.5340 \\ \hline
	
	\multirow{2}{*}{SPMLD}
	
	&30\%  &\textbf{0.104 $\pm$ 0.0030} & \textbf{0.903 $\pm$ 0.0047} & \textbf{4.600 $\pm$ 0.5359}  \\
	&70\%  &\textbf{0.113 $\pm$ 0.0015} & \textit{0.894 $\pm$ 0.0053} &  \textit{5.018 $\pm$ 0.5712} \\
		\hline
	\end{tabular*}
\end{table}
\begin{table}

	\caption{Average performance on \textit{Education} dataset: means and standard deviations over 10 independent runs. The best performance is highlighted in \textbf{bold} with respect to paired t-tests at 95\% significance level between SPMLD and each baseline.}
	\begin{tabular*}{\textwidth}{@{\extracolsep{\fill} }lccccc}  
		\hline
		Method &$\rho$   & Rkl $\downarrow$& AUC $\uparrow$ &COV $\downarrow$  \\ 
		\hline
		\multirow{2}{*}{LEML}
		
		&30\%  & 0.176 $\pm$ 0.0084 & 0.817 $\pm$ 0.0075 & 9.672 $\pm$ 0.4461  \\
		&70\%  & 0.151 $\pm$ 0.0077 & 0.842 $\pm$ 0.0082 & 7.595 $\pm$ 0.5138  \\ \hline
		
		\multirow{2}{*}{ML-LRC}
		
		&30\%  & 0.144 $\pm$ 0.0034 & 0.845 $\pm$ 0.0068 & 6.350 $\pm$ 0.2042  \\
		&70\%  & 0.113 $\pm$ 0.0028 & 0.860 $\pm$ 0.0063 & 5.075 $\pm$ 0.1866  \\ \hline
		
		\multirow{2}{*}{GLOCAL}
		
		&30\%  & 0.125 $\pm$ 0.0026 & 0.875 $\pm$ 0.0057 & 5.741 $\pm$ 0.2312  \\
		&70\%  & 0.122 $\pm$ 0.0035 & 0.878 $\pm$ 0.0064 & 5.784 $\pm$ 0.2450 \\ \hline
		
		\multirow{2}{*}{SPMLD}
		
		&30\%  & \textbf{0.096 $\pm$ 0.0019} & \textbf{0.904 $\pm$ 0.0048} & \textbf{4.246 $\pm$ 0.1372}   \\
		&70\%  & \textbf{0.093 $\pm$ 0.0021} & \textbf{0.907 $\pm$ 0.0052} & \textbf{4.162 $\pm$ 0.1524}   \\ 
		\hline
	\end{tabular*}
\end{table}

\begin{table}

	\caption{Average performance on \textit{Health} dataset: means and standard deviations over 10 independent runs. The best performance is highlighted in \textbf{bold} with respect to paired t-tests at 95\% significance level between SPMLD and each baseline.}
	\begin{tabular*}{\textwidth}{@{\extracolsep{\fill} }lccccc}  
		\hline
		Method &$\rho$   & Rkl $\downarrow$& AUC $\uparrow$ &COV $\downarrow$  \\ 
		\hline
		\multirow{2}{*}{LEML}
		&30\%  & 0.095 $\pm$ 0.0029 & 0.896 $\pm$ 0.0038 & 6.248 $\pm$ 0.1640  \\
		&70\%  & 0.074 $\pm$ 0.0033 & 0.920 $\pm$ 0.0045 & 5.167 $\pm$ 0.1827  \\ \hline
		\multirow{2}{*}{ML-LRC}
		&30\%  & 0.085 $\pm$ 0.0041 & 0.907 $\pm$ 0.0082 & 4.924 $\pm$ 0.1351  \\
		&70\%  & 0.071 $\pm$ 0.0036 & 0.920 $\pm$ 0.0093 & 3.960 $\pm$ 0.1825  \\ \hline
		\multirow{2}{*}{GLOCAL}
		&30\%  & 0.0828 $\pm$ 0.0018 & 0.919 $\pm$ 0.0057 & 4.438 $\pm$ 0.1357\\
		&70\%  & 0.0795 $\pm$ 0.0023 & 0.923 $\pm$ 0.0078 & 4.472 $\pm$ 0.1414\\ \hline
		\multirow{2}{*}{SPMLD}
		&30\%  &\textbf{ 0.064 $\pm$ 0.0009}  & \textbf{0.938 $\pm$ 0.0063}  &\textbf{ 3.355 $\pm$ 0.1182}\\
		&70\%  &\textbf{ 0.059 $\pm$ 0.0011}  & \textbf{0.943 $\pm$ 0.0068}  &\textbf{ 3.253 $\pm$ 0.1200}\\ 
		\hline
	\end{tabular*}
\end{table} 

\begin{table}
	\caption{Average performance on \textit{Science} dataset: means and standard deviations over 10 independent runs. The best performance is highlighted in \textbf{bold} with respect to paired t-tests at 95\% significance level between SPMLD and each baseline.}
	\begin{tabular*}{\textwidth}{@{\extracolsep{\fill} }lccccc}  
		\hline
		Method &$\rho$   & Rkl $\downarrow$& AUC $\uparrow$ &COV $\downarrow$  \\ 
		\hline
		\multirow{2}{*}{LEML}
		&30\%  & 0.203 $\pm$ 0.0052  & 0.827 $\pm$ 0.0053 & 10.587 $\pm$ 0.2011  \\
		&70\%  & 0.174 $\pm$ 0.0056  & 0.849 $\pm$ 0.0064 & 9.501 $\pm$ 0.2548 \\ \hline
		\multirow{2}{*}{ML-LRC}
		&30\%  & 0.169 $\pm$ 0.0027 & 0.830 $\pm$ 0.0039 & 8.794 $\pm$ 0.1254  \\
		&70\%  & 0.134 $\pm$ 0.0024 & 0.850 $\pm$ 0.0033 & 6.900 $\pm$ 0.1273 \\ \hline
		\multirow{2}{*}{GLOCAL}
		&30\%  & 0.154 $\pm$ 0.0029 & 0.840 $\pm$ 0.0108 & 7.949 $\pm$ 0.1371  \\
		&70\%  & 0.134 $\pm$ 0.0034 & 0.866 $\pm$ 0.0113 & 7.106 $\pm$ 0.1365 \\ \hline
		\multirow{2}{*}{SPMLD}
		&30\%  &\textbf{0.129 $\pm$ 0.0024}  & \textbf{0.871 $\pm$ 0.0095} & \textbf{6.640 $\pm$ 0.1156}    \\
		&70\%  &\textbf{0.124 $\pm$ 0.0028}  & \textbf{0.876 $\pm$ 0.0087} & \textbf{6.432 $\pm$ 0.1283}   \\ 
		\hline
	\end{tabular*}
\end{table}

\begin{table}

	\caption{Average performance on \textit{Social} dataset: means and standard deviations over
		10 independent runs. The best performance is highlighted in \textbf{bold} with respect to paired t-tests at 95\% significance level between SPMLD and each baseline. \textit{Italic} font indicates that SPMLD and the corresponding baselines show no significant difference.}
	\begin{tabular*}{\textwidth}{@{\extracolsep{\fill} }lccccc}  
		\hline
		Method &$\rho$   & Rkl $\downarrow$& AUC $\uparrow$ &COV $\downarrow$  \\ 
		\hline
		\multirow{2}{*}{LEML}
		&30\%  & 0.128 $\pm$ 0.0067 & 0.872 $\pm$ 0.0065 & 5.459 $\pm$ 0.3084   \\
		&70\%  & 0.081 $\pm$ 0.0061 & 0.919 $\pm$ 0.0069 & 3.824 $\pm$ 0.3011  \\ \hline
		\multirow{2}{*}{ML-LRC}
		&30\%  & 0.123 $\pm$ 0.0059 & 0.877 $\pm$ 0.0057 & 5.167 $\pm$ 0.1034   \\
		&70\%  & 0.073 $\pm$ 0.0052 & 0.928 $\pm$ 0.0046 & 3.608 $\pm$ 0.0975  \\ \hline
		\multirow{2}{*}{GLOCAL}
		&30\%  & 0.102 $\pm$ 0.0054 & 0.898 $\pm$ 0.0058 & 4.496 $\pm$ 0.2627  \\
		&70\%  & 0.073 $\pm$ 0.0049 & 0.929 $\pm$ 0.0052 & \textit{3.442 $\pm$ 0.2583}   \\ \hline
		\multirow{2}{*}{SPMLD}
		&30\%  &\textbf{ 0.068 $\pm$ 0.0055}  & \textbf{0.931 $\pm$ 0.0052} & \textbf{3.563 $\pm$ 0.2554}   \\
		&70\%  &\textbf{ 0.065 $\pm$ 0.0048}  & \textbf{0.934 $\pm$ 0.0044}  & \textit{3.456 $\pm$ 0.2505} \\ 
		\hline
		
	\end{tabular*}
		\end{table}
\begin{figure}
	\includegraphics[width=\textwidth]{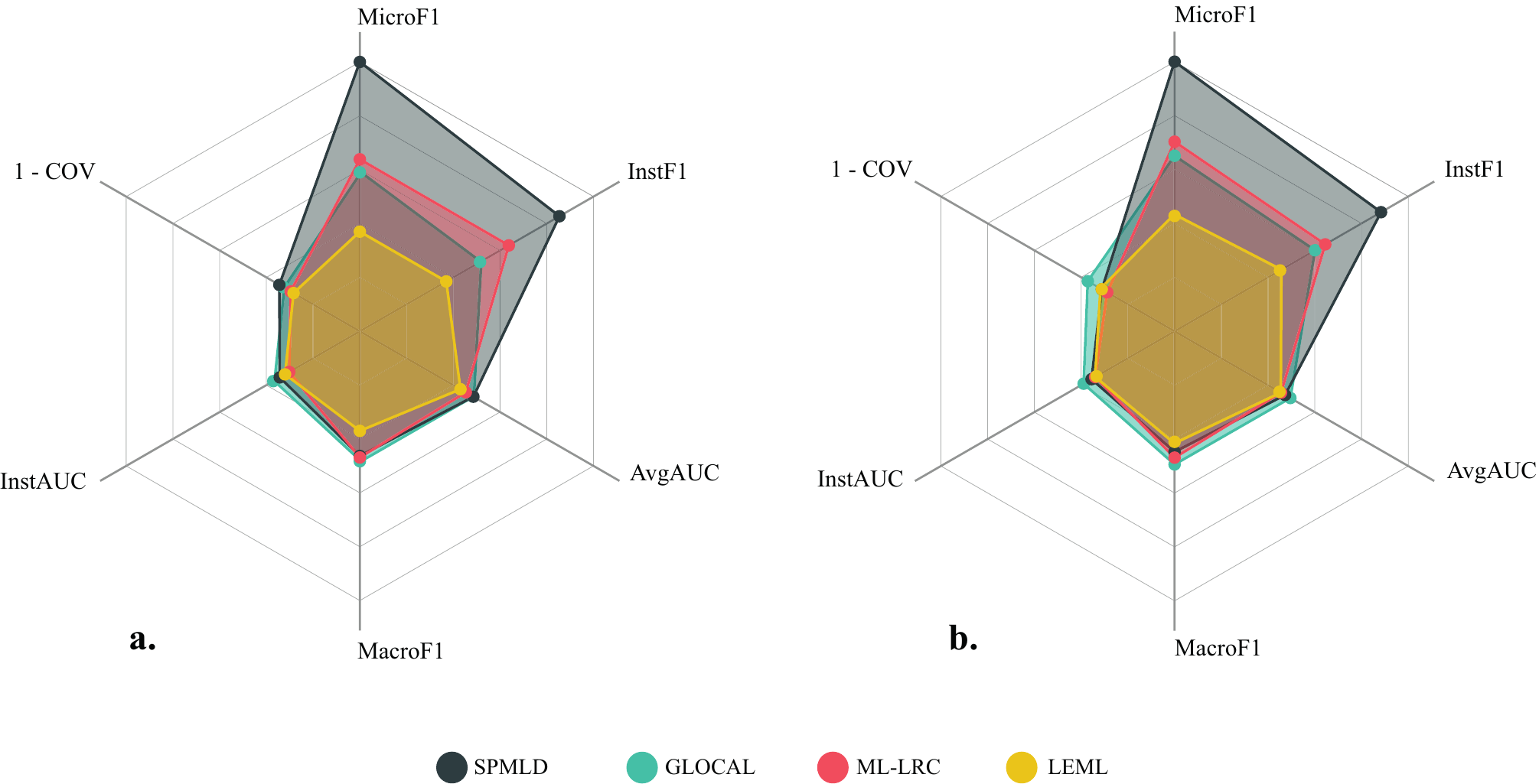}
	\caption{The comparison on
		Corel5K dataset with respect to several metrics (the coverage measure is normalized). a) 30\% label rate , b) 70\% label rate.} \label{fig1}
\end{figure}
Results obtained on \textit{Corel5K} image dataset are analyzed through a Radar plot which enables one to compare models against multiple metrics. In this straightforward analysis, SPMLD is compared to the baselines in the high-dimensional label space of \textit{Corel5K} in terms of six evaluation metrics for 30\% and 70\% revealed data, respectively and the results are shown in Figure~\ref{fig1}. Note that, the proposed method endeavors to cover all labels fairly in its learning process. Thus, it is able to distinguish positive and negative labels of an instance by simultaneously making a larger label-wise margin and preserving instance-wise margin. Hence, it can be obviously seen that the SPMLD is far better on label-wise metrics (MicroF1 and InstanceF1) and it obtains comparable results on the other instance-wise metrics such as MacroF1. 

\begin{figure}
	\includegraphics[width=\textwidth]{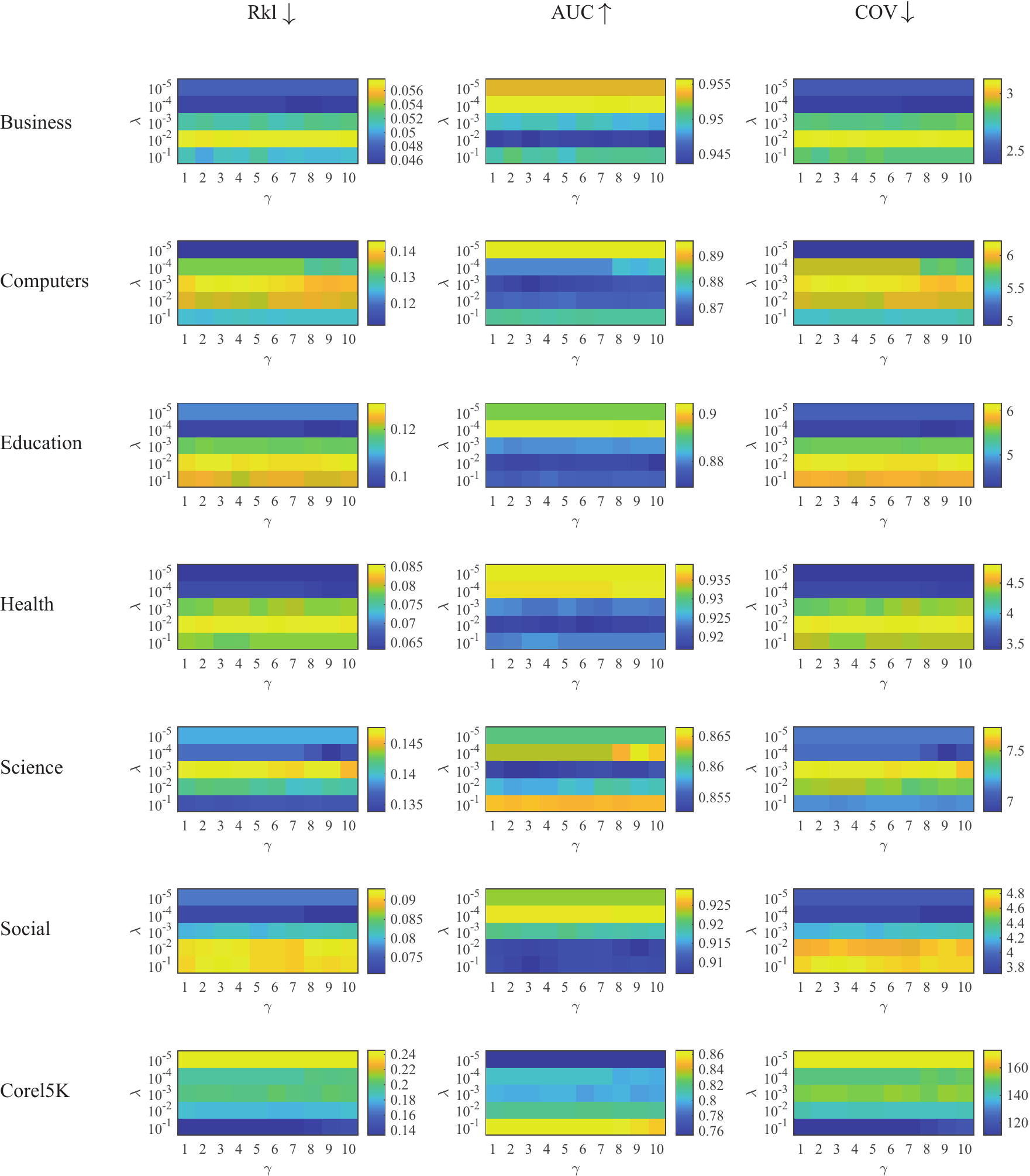}
	\caption{Analysis of influence of $\lambda$ and $\gamma$ on SPMLD for 30\% data revealed.} \label{fig2}
\end{figure}
\subsection{Parameter Analysis}

In this subsection, the influence of parameters on the proposed model is analyzed. According to (\ref{eq8}) SPMLD has two parameters namely $\lambda$ and $\gamma$ which correspond to the self-paced and the diversity regularization terms, respectively. It must be mentioned that parameters of other regularization terms which belong to the base algorithm are analyzed in \cite{8233207}. Thus, to make a thorough study on the two mentioned parameters we evaluated them on all datasets through a grid-search strategy and reported the results based on three measures. In addition, each graph consists of 50 evaluations referring to 5 different $\lambda$s and 10 $\gamma$s. According to Figure~\ref{fig2}, there is a bar next to each graph that indicates the highest and the lowest values obtained on the corresponding dataset regarding each measure which is shown using an spectrum of colors. Subsequently, Light colors (e.g. "orange" to "yellow") represent high amounts and dark colors (e.g. "blue" to "dark-blue") represent low amounts for the measures.\\
It can be seen that for each value of $\lambda$ changing the values of $\gamma$ from 1 to 10 makes a significant difference except for $\lambda$=$10^{-5}$ which lies on the top row of graphs and stays unchanged with increasing $\gamma$.

\section{Conclusion}
In this paper, we propose a novel Self-Paced framework for Multi-Label learning. This framework incorporates the complexity of both instances and labels, and trains its predictive model with gradual involvement of harder samples. It also utilizes an efficient Diversity maintenance mechanism to avoid biasing over a limited subset of labels. The diverse easy-to-hard learning strategy has also an implicit positive effect on correlations exploited. SPMLD is applied to correlation-based multi-label learning as a host algorithm. Experiments on real-world datasets verify the effectiveness of SPMLD compared to the host algorithm itself and two other state-of-the-art methods. For future studies, it is desirable to investigate the direct impact of self-paced regularization on correlation exploitation and we intend to study and analyze the effect of diversity on local dependencies.

\bibliography{seyedi19a}
\end{document}